\documentclass[runningheads]{llncs}
\usepackage{graphicx}
\usepackage{amsmath,amssymb} 
\usepackage{color}

\begin{document}
\pagestyle{headings}
\mainmatter

\def\ACCV20SubNumber{802}  

\title{ERIC: Extracting Relations Inferred from Convolutions} 
\titlerunning{ERIC: Extracting Relations Inferred from Convolutions}
%
\author{Joe Townsend\inst{1}\orcidID{0000-0002-5478-0028} \and
Theodoros Kasioumis\inst{1}\orcidID{0000-0003-2008-5817} \and
Hiroya Inakoshi\inst{1}\orcidID{0000-0003-4405-8952}}
\authorrunning{J. Townsend et al.}
%
\institute{Fujitsu Laboratories of Europe LTD, 4th Floor, Building 3, Hyde Park Hayes, 11 Millington Road, Hayes, Middlesex, UB3 4AZ, United Kingdom
\email{\{joseph.townsend,theodoros.kasioumis,hiroya.inakoshi\}@uk.fujitsu.com}}

\maketitle

\begin{abstract}
Our main contribution is to show that the behaviour of kernels across multiple layers of a convolutional neural network can be approximated using a logic program. The extracted logic programs yield accuracies that correlate with those of the original model, though with some information loss in particular as approximations of multiple layers are chained together or as lower layers are quantised. We also show that an extracted program can be used as a framework for further understanding the behaviour of CNNs. Specifically, it can be used to identify key kernels worthy of deeper inspection and also identify relationships with other kernels in the form of the logical rules. Finally, we make a preliminary, qualitative assessment of  rules we extract from the last convolutional layer and show that kernels identified are symbolic in that they react strongly to sets of similar images that effectively divide output classes into sub-classes with distinct characteristics.
\end{abstract}

\section{Introduction}
\label{sect:introduction}

As public concern regarding the extent to which artificial intelligence can be trusted increases, so does the demand for so-called \emph{explainable} AI. While accountability is a key motivator in recent years, other motivations include understanding how models may be improved, knowledge discovery through the extraction of concepts learned by the models but previously unknown to domain experts, means by which to test models of human cognition, and perhaps others. 

This has led to extensive research into explaining how models trained through machine learning make their decisions \cite{ribeiro2016should,gilpin2018explaining,guidotti2018survey}, and the field of \emph{Neural-Symbolic Integration} covers this work with respect to neural networks \cite{andrews1995survey,jacobsson2005rule,townsend2019extracting,zhang2018visual,lamb2020graph}. The latter began by largely focussing on modelling the behaviour of multi-layer perceptrons or recurrent neural networks as symbolic rules that describe strongly-weighted relationships between neurons in adjacent layers \cite{andrews1995survey,jacobsson2005rule}. More recent work strives to explain deeper networks, including convolutional neural networks (CNNs) \cite{jacobsson2005rule,zhang2018visual,townsend2019extracting,lamb2020graph}. Most of these methods identify important input or hidden features with respect to a given class or convolutional kernel \cite{simonyan2013deep,zeiler2014visualizing,springenberg2014striving,bojarski2016visualbackprop,bach2015pixel,samek2017evaluating,shrikumar2017learning}, but methods that extract rule or graph-based relationships between key features are also emerging \cite{frosst2017distilling,chen2019looks,bologna2020two,zhang2017growing,zhang2018interpreting,zhang2019interpreting}. Moreover it has been shown that a CNN's kernels may correspond to semantically meaningful concepts to which we can ascribe symbols or words \cite{zhou2014object}.

We show how the behaviour of a CNN can be approximated by a set of logical rules in which each rule's conditions map to convolutional kernels and therefore the semantic concepts they represent. We introduce \emph{ERIC (Extracting Relations Inferred from Convolutions)}, which assumes each kernel maps to an individual concept, quantises the output of each kernel as a binary value, extracts rules that relate the binarised kernels to each other and visualises the concepts they represent. We also argue that the extracted rules simplify the task of identifying key kernels for inspection (using for example importance methods described above), as the number of kernels in a layer is often in the order of hundreds.

Although related work which extracts graph-based approximations has also made significant strides in this direction \cite{zhang2017growing,zhang2018interpreting}, so far nodes in the graph only correspond to positive instances of symbols, e.g. ``If feature X is observed...'', and not negative, e.g. ``If X is not observed...''. Propositional logic is able to express both ($X$ and $\lnot X$). Furthermore our method is entirely post-hoc and does not assume a convolutional architecture has been designed  \cite{chen2019looks,bologna2020two} or trained \cite{zhang2018interpretable} to learn semantically meaningful kernels. However ERIC is not necessarily incompatible with such architectures either, allowing for flexible usage.

We begin with a literature survey in section \ref{sect:background}, and section \ref{sect:architecture} outlines ERIC's architecture. In section \ref{sect:experiments} we extract logic programs from multiple convolutional layers and show that these programs can approximate the behaviour of the original CNN to varying degrees of accuracy depending on which and how many layers are quantised. Section \ref{sect:experiments} ends with an analysis of extracted rules and argues that the kernels they represent correspond to semantically meaningful concepts. The discussion in section \ref{sect:discussion} argues that the extracted rules faithfully represent how the CNN `thinks', compares ERIC to other methods from the literature and also proposes future work. Section \ref{sect:conclusion} presents our conclusion that kernels can be mapped to symbols that, regardless of their labels, can be manipulated by a logic program able to approximate the behaviour of the original CNN.

\section{Background}
\label{sect:background}

\subsection{Rule extraction from neural networks}
\label{sect:bgrexnn}

Since at least the 1990s efforts have been made to extract interpretable knowledge from neural networks, and during this period Andrews et al. defined three classes of extraction method \cite{andrews1995survey}. \emph{Pedagogical} methods treat a network as a black box and construct rules that explain the outputs in terms of the inputs. \emph{Decompositional} methods extract separate rule sets for individual network parts (such as individual neurons) so that collectively all rules explain the behaviour of the whole model. \emph{Eclectic} methods exhibit elements of both of the other classes.

Another important distinction between different classes of extraction method is the \emph{locality} of an explanation \cite{percy2016need,ribeiro2016should}. Some extraction methods provide \emph{local} explanations that describe individual classifications, wheras others are more \emph{global} in that they provide explanations for the model as a whole.

Two important components for extracting rules from a network are \emph{quantisation} and \emph{rule construction} \cite{jacobsson2005rule,townsend2019extracting}. \emph{Quantisation} maps input, hidden and output states of neural networks from the domain of real numbers to binary or categorical values, for example by thresholding. \emph{Rule construction} forms the rules which describe the conditions under which these quantised variables take different values (e.g. true or false) based on the values of other quantised variables.

In addition to measuring classification accuracy of an explainable approximation of a model, it is also common to record \emph{fidelity} to the behaviour of the original model. In other words, \emph{fidelity} is the accuracy of the approximation with respect to the outputs of the original model. Also, if a model is to be regarded as `explainable', then there must be some means by which to quantify this quality. Explainability is a subjective quality and at the time of writing there does not appear to be a consensus on how to quantify it. Examples of the various approaches include counting extracted rules \cite{andrews1995survey} or some assessment of how humans respond to or interact with extracted rules presented to them \cite{percy2016need,ribeiro2018anchors}.

However explainability is quantified, it is often observed that there is a trade-off between an extraction method's explainability and its fidelity due to information loss that results from quantifying continuous variables. The preference of fidelity and accuracy over explainability or vice-versa may depend on the nature of the task or a user's preference \cite{percy2016need}. If the model is advising a human decision-maker such as a doctor who has to justify their decisions to others, then explainability is key. For a task that is entirely automated but not safety-critical to the extent that such accountability is required, then explainability can be sacrificed for accuracy. That said, in the latter case, \emph{some} explainability is still useful as humans may discover new knowledge by analysing what the automated system has learned. In situations where accountability is a priority, one may prefer network architectures that are themselves designed or trained with explainability in mind. Solutions like these are often described as \emph{explainable-by-design} and for brevity we abbreviate these to \emph{XBD-methods}. However in XBD methods it may be more difficult to discover new knowledge as they explore a more constrained search space during training.

Early work largely focussed on multi-layer perceptrons (MLPs) with one or very few hidden layers and also on recurrent neural networks. Research has since grown into explaining `deeper' neural networks of several to many layers, be these MLPs that are deep in this particular sense \cite{zilke2016deepred,schaaf2019enhancing,nguyen2020towards} or more advanced architectures such as LSTMs \cite{murdoch2017automatic}, Deep Belief Networks \cite{tran2016deep} or CNNs \cite{frosst2017distilling,chen2019looks,bologna2020two,zhang2017growing,zhang2018interpreting,zhang2019interpreting}. Remaining subsections only cover methods that extract explanations from CNNs. The reader is referred to surveys in the literature regarding other network types \cite{jacobsson2005rule,zhang2018visual,townsend2019extracting,lamb2020graph}. We also acknowledge generic methods that treat arbitrary models as black boxes but do not cover them as they are \emph{pedagogical} and by nature cannot decompose neural networks. These are also surveyed in the literature \cite{gilpin2018explaining,guidotti2018survey}.

\subsection{Feature importance}
\label{sect:bgfeatures}

A lot of existing research presents ways to visualise what CNNs `see' \cite{simonyan2013deep,zeiler2014visualizing,springenberg2014striving,bojarski2016visualbackprop,bach2015pixel,samek2017evaluating,shrikumar2017learning}. These methods generally identify the responsibility of input pixels (or neurons in a hidden layer) with respect to activating the output neuron corresponding to a given class. This usually involves tracing the signal back from that output neuron, backwards through the network along stronger network weights until arriving at the input image. This signal may be the output activation \cite{zeiler2014visualizing,zhou2016learning}, a gradient \cite{simonyan2013deep,denil2014extraction,selvaraju2017grad} or some other metric derived from the output \cite{bach2015pixel,samek2017evaluating}. These ideas can be used to analyse what a specific kernel responds to \cite{zeiler2014visualizing}. Furthermore, Zhou et al. show that semantic concepts can be observed from an analysis of a kernel's receptive field in CNNs trained to recognise scenes, and that kernels tend to have more semantic meaning at deeper network layers \cite{zhou2014object}. In related work Simon et al provide a means of localising semantic parts of images \cite{simon2014part}.

\subsection{Rule extraction from CNNs}
\label{sect:bgrexcnn}

Compared with methods for visualising important features as in section \ref{sect:bgfeatures}, methods that model the relationships between these features are relatively few. 

Chen et al. introduce an XBD-model that includes a \emph{prototype} layer that is trained to recognise a set of prototype components so that images can be classified by reference to these component parts. In other words, the CNN is trained to classify images in a human-like manner. For example, one kernel learns the concept of wing, another learns the concept of beak, and when an input image is classified the explanation can be given as $wing \land beak \rightarrow bird$. 

The prototype method, and currently our own, assumes a one-to-one relationship between kernels and symbols. However it has been observed that this may not be the case \cite{xie2017relating}. It may be that the relationship between kernels and semantic concepts is in fact many-to-many. Zhang et al. disentangle concepts represented in this way and represent disentangled concepts and their relationship to each other in a hierarchical graph in which each layer of the hierarchy corresponds to a layer of the CNN \cite{zhang2017growing,zhang2018interpreting}. However, the disentangled graphs in their current form show limited expressivity in that explanations are only composed of positive instances of parts. We extract rules in which conditions may be positive or negative. The work was extended to an XBD approach in which a CNN is trained with a loss function that encourages kernels to learn disentangled relations \cite{zhang2018interpretable}, and this was then used to generate a decision tree based on disentangled parts learned in the top convolutional layer \cite{zhang2019interpreting}.

Bologna and Fossati extract propositional rules from CNNs \cite{bologna2020two}. First they extract rules that approximate the dense layers, with antecedents corresponding to the outputs of individual neurons in the last convolutional layer, and then extract rules that summarise the convolutional layers, with antecedents mapped to input neurons. This work is to some extent XBD as it assumes that some layers of the original model are discretised. Only the dense layer rules are actually used for inference, with convolutional rules only used to provide explanations. The complexity of working with invdidual neurons as antecedents is cited as the reason for this decision. Other work described above (and ours) overcomes this by mapping symbols to \emph{groups} of neurons (e.g. prototype kernels or disentangled parts). One advantage over the disentanglement method is that extracted rules may include negated antecedents.


\section{ERIC Architecture}
\label{sect:architecture}

\begin{figure}[t]
\centering
\includegraphics[width=120mm]{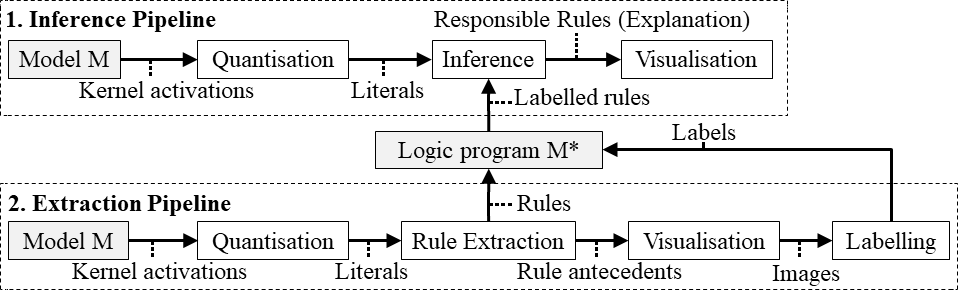} 
\caption{ERIC Pipelines for \emph{inference} and \emph{rule extraction}.}
\label{fig:pipeline}
\end{figure}

ERIC is a global explanation method that extracts rules conditioned on positive and negative instances of quantised kernel activations and is able to extract these rules from multiple convolutional layers. ERIC assumes CNNs have standard convolution, pooling and dense layers, and is indifferent with respect to whether the CNN has been trained with explainability in mind. ERIC is mostly decompositional in that rules explain kernel activations but partly pedagogical in that we only decompose a subset of convolutional layers and the output dense layer, and treat remaining layers as black boxes. Fig. \ref{fig:pipeline} presents an overview of the architecture as two pipelines sharing most modules. We explain the inference module first in section \ref{sect:inference} in order to formalise the target behaviour of extracted programs. All modules of the extraction pipeline are explained in section \ref{sect:extraction}.

\subsection{Preliminaries}

Let us consider a set of input images $\boldsymbol{x}$ indexed by $i$ and a CNN $M$ whose layers are indexed by $l = 1, \dots, l^o$. Every layer has kernels indexed by $k = 1, \dots, K_l$. $A_{i,l,k} \in \mathbb{R}^{h \times w}$ denotes an activation matrix output for a kernel, where $h,w$ are natural numbers. Note that we treat the term \emph{kernel} as synonymous with \emph{filter} and we do not need to consider a kernel's input weights for our purposes in this paper. Let $o$ refer to the softmax layer at the output of $M$, with index $l^o$. Let $l^{LEP}$ denote the index of a special layer we call the \emph{Logical Entry Point}, the layer after which and including we approximate kernel activations. 

Let $b_{i,l,k} \in \{1,-1\}$ denote a binary truth value associated with $A_{i,l,k}$ as in eq. \ref{eq:infer} and \ref{eq:infer2}. $b_{i,l,k}$ may be expressed as positive and negative literals $\mathcal{L}_{i,l,k} \equiv (b_{i,l,k} = 1)$ and $\lnot \mathcal{L}_{i,l,k} \equiv (b_{i,l,k} = -1)$ respectively. A set of rules indexed by $r$ at layer $l$ is denoted $R_{l} = \{ R_{l,r} = (D_{l,r},C_{l,r}) \}_{r}$, where $D_{l,r}$ and $C_{l,r}$ are sets of conjoined literals in the \emph{antecedents} (conditions for satisfying the rule) and \emph{consequents} (outcomes) of $R_{l,r}$ respectively. For example, $D_{l,r} = \mathcal{L}_{i,l-1,3} \wedge \lnot \mathcal{L}_{i,l-1,6} \wedge \mathcal{L}_{i,l-1,7}$ and $C_{l,r} = \mathcal{L}_{i,l,2} \wedge \mathcal{L}_{i,l,3} \wedge \mathcal{L}_{i,l,5}$. $C_{l,r}$ may only contain positive literals as we assume \emph{default negation}, i.e. by default all $b_{i,l,k} = -1$ ($\lnot \mathcal{L}_{i,l,k}$) unless some $C_{l,r}$ determines otherwise. 

\subsection{Inference}
\label{sect:inference}

\begin{figure}[t]
\centering
\includegraphics[width=120mm]{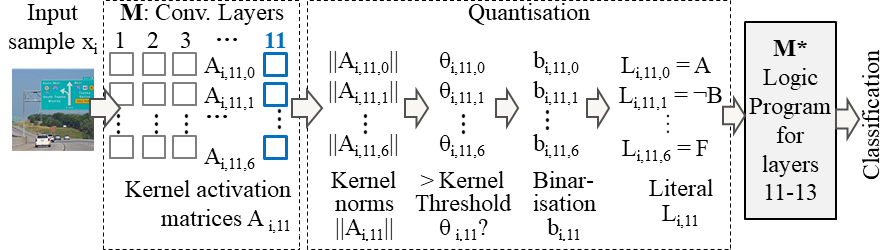} 
\caption{Inference: kernel outputs at a designated layer are \emph{quantised} and input to a \emph{logic program} that approximates remaining layers. \emph{Input sample} from Places365 \cite{zhou2017places}.}
\label{fig:inference}
\end{figure}

Inference is summarised in fig. \ref{fig:inference}. Eq. \ref{eq:infer} and \ref{eq:infer2} formalise the process by which we infer a binary approximation $b_{i,l,k}$ for activation tensor $A_{i,l,k}$ for any kernel. An extracted program approximates convolutional layers after and including layer $l^{LEP}$, at which point kernel activations are mapped to binary values via a quantisation function  $Q(A_{i,l,k}, \theta_{l,k})$ so that these activations may be treated as the input to logic program $M^*$ (eq. \ref{eq:infer}).  $Q(A_{i,l,k}, \theta_{l,k})$ is explained in detail later in subsection \ref{sect:quantisation}. The truths of all kernels in all following layers ($b_{i,l,k}$ for $l > l^{LEP}$) are derived through logical inference on the truths of binarised kernels from the previous layer $\boldsymbol{b}_{i,l-1}$ according to a set of layer-specific rules $R_l$ (eq. \ref{eq:infer2}).

\begin{equation} \label{eq:infer}
	b_{i,l^{LEP},k} = Q(A_{i,l^{LEP},k}, \theta_{l^{LEP},k})
\end{equation}

\begin{equation} \label{eq:infer2}
	b_{i,l,k} = \begin{cases}
				1 \text{ depending on } C_{l,r} \text{ for all } k \text{ if } \exists r (D_{l,r} = True) \\ 
				-1 \text{ otherwise (default negation)}
			\end{cases}
\end{equation}

\subsection{Rule extraction}
\label{sect:extraction}

Rule extraction is implemented as a pipeline of 5 modules (fig. \ref{fig:pipeline}). First is the original \textbf{model} $M$ for which we want to extract an approximation $M^*$. We do not need to say much about this except that ERIC assumes $M$ has already been trained. Next in the \textbf{quantisation} stage we obtain binarisations for all kernels after and including layer $l^{LEP}$ based on activations obtained for training data. We then \textbf{extract rules} which describe the relationship between kernels by reference to their binarisations. Then to interpret the meanings of individual kernels we first \textbf{visualise} each kernel as one or more images that represent what inputs the kernels strongly react to, before each kernel is assigned a \textbf{label} based on manual inspection, a process we plan to automate in future work.

\subsubsection{Quantisation}
\label{sect:quantisation}

Our quantisation function $Q$ is defined in eq. \ref{eq:quant}, where $\theta_{l,k}$ is a kernel-specific threshold and norm function $\|\cdot\|$ is the \emph{l1} norm\footnote{Preliminary experiments found that \emph{l1} norm yielded higher fidelity than \emph{l2} norm.}. Intuitively, we say that a kernel is active when its norm breaches a threshold specific to that kernel. Note that for the initial rule extraction process we quantise all extractable layers $l \geq l^{LEP}$ but for inference we only need to quantise kernels at $l^{LEP}$.

\begin{equation} \label{eq:quant}
	Q(A_{i,l,k}, \theta_{l,k}) = \begin{cases}
								1 \text{ if } \|A_{i,l,k}\| > \theta_{l,k}\\
								-1 \text{ otherwise}
							\end{cases}
\end{equation}

We define a kernel's threshold as the mean norm of its activations with respect to the training data $x^{tr}$, as in eq \ref{eq:threshset}. To this end we make a forward pass of $x^{tr}$ in order to obtain $\{ A^{\text{tr}}_{i,l,k} \}_{i,l,k} $, activations for each kernel $1 \leq k \leq K_l$ at each layer $ l^{LEP} \leq l < l^o$ for each input training sample $1 \leq i \leq n$. 

\begin{equation} \label{eq:threshset}
	\theta_{l,k} = \frac{\sum\nolimits^n_{i=1}{\|A^{tr}_{i,l,k}\|}}{n}
\end{equation}

We can now use the quantisation function (eq. \ref{eq:quant}) to obtain binarisations of all kernel activations according to eq. \ref{eq:trbin}. Where a convolutional layer outputs to a pooling layer, we take $A_{i,l,k}$ from the pooled output. As also shown in eq. \ref{eq:trbin}, we also need to treat output neurons as kernels of dimension $1 \times 1$ so that $\boldsymbol{b}^{tr}_{i,l^o} = M(x^{tr}_i)$. This enables us to extract rules that map kernel activations at layer $l^o-1$ to the output classifications as inferred by $M$.

\begin{equation} \label{eq:trbin}
	b_{i,l,k} = \begin{cases}
				o^{tr}_{i,k} \text{ if } l = l^o \\ 
				Q(A^{tr}_{i,l,k}) \text{ otherwise} 
			\end{cases}
\end{equation}

\subsubsection{Rule extraction}

We now extract rules that describe the activation at each kernel at every layer $l$ given activations at layer $l-1$. Thus, the following is applied layer-wise from $l^o$ to $l^{LEP}$. We use a tree-based extraction algorithm similar to the C4.5 algorithm \cite{quinlan1993c4} to extract rules which describe conditions for which each kernel evaluates as true. As we assume default negation, we do not need to extract rules that describe when a kernel is false. Let us denote the training data $Z_l =\{ (\boldsymbol{z}_i, t_i) \mid i=1,...,n \}$ where $\boldsymbol{z}_i \in \{True,False\}^{2K_{l-1}}$ and $t_i \in \{True,False\}$. Note that  the length of $\boldsymbol{z}_{l}$ is twice the number of kernels at layer $l-1$ because each kernel has positive and negative literals. $z_{l-1,k'} = True$ if it corresponds to a positive literal and its binary value is 1 or if it represents a negative literal and its binary value is -1. It is False otherwise.  C4.5 generates a decision tree up to maximum depth $d$. Each path from the root of the tree to a leaf node represents a separate rule and nodes branch on rule conditions (i.e. antecedents). The maximum number of antecedents per rule is equal to $d+1$. C4.5 uses entropy as a branch selection criterion but based on a marginal improvement in fidelity observed in preliminary tests we chose to use gini index. Extraction can become intractable as more layers and therefore kernels are introduced due to combinatorial explosion. This can be moderated by reducing $d$ or increasing another parameter $\alpha$ that we introduce for this purpose. Let P, Q represent sets of training instances that satisfy the path of conditions leading to a parent node and child node, respectively. We stop branching if $|Q| / |P| < \alpha$. If a leaf node represents multiple outcomes, we set the consequence to the modal value of $Q$.  Finally, we simplify extracted programs by merging rules with complementary literals ($A \land B \rightarrow C$ and $A \land \lnot B \rightarrow C$ become $A \rightarrow C$) and rules with identical antecedents but different consequents ($A \rightarrow B$ and $A \rightarrow C$ become $A \rightarrow B \land C$). 

\subsubsection{Kernel visualisation and labelling}
\label{sect:arch_kvis}

To visualise what a kernel sees, we select the $m$ images from $x^{tr}$ which activate that kernel most strongly with respect to $\|A^{tr}_{i,l,k}\|$. We denote this visualisation as $\hat{x}^{m}_{l,k}$. A label is assigned to a kernel based on $\hat{x}^{m}_{l,k}$, which for the time being we do manually based on visual inspection but in future work plan to automate. For the time being, to defend the arguments of the paper, it is not so much the labels that are important as the distinction between the subsets of image that each kernel responds most strongly to.

\section{Experiments}
\label{sect:experiments}

In sections \ref{sect:exp_task} and \ref{sect:exp_model} we outline the classification task and CNN configuration we use for our experiments. We then extract rules from a single convolutional layer in section \ref{sect:exp_single} and then multiple convolutional layers in section \ref{sect:exp_multi}. In section \ref{sect:exp_vislab} we visualise and label kernels and in section \ref{sect:exp_rules} analyse some of the rules with these labels assigned to the antecedents.

\subsection{Task}
\label{sect:exp_task}

We chose to classify road scenes from the places365 dataset \cite{zhou2017places} for a number of reasons. First, we felt that a scene dataset was appropriate as scenes can easily be described by reference to symbolic entities within them which themselves could be described by a separate classifer (i.e. the kernel classifier) with a large vocabulary of labels. We selected a handful of 5 scenes to simplify the task given the complexity of rule extraction, and opted for roads in order to create a scenario where the distinction between scenes is particularly important (due to regulations, potential hazards, etc). We wanted to demonstrate ERIC on a multi-class task and on multiple combinations of class. 3 is the minimum required for multi-class case and gives us ${5 \choose 3} =10$ combinations of scenes (table \ref{table:netcompare}).

\subsection{Network Model}
\label{sect:exp_model}

For each combination of classes we train VGG16 (as defined for Tensorflow using Keras \cite{chollet2015keras}) from scratch over 100 epochs using Stochastic Gradient Descent, a learning rate of $10^{-5}$, categorical crossentropy and a batch size of 32.

\subsection{Extraction from a single layer}
\label{sect:exp_single}

\begin{table}[t]
	\caption{\emph{Accuracies} of the \emph{original model} $M$ and \emph{extracted program} $M^*$, with the number of unique \emph{variables} (positive or negative literals) and \emph{rules} for each $M^*$, and the \emph{size} of $M^*$ measured as the total number of antecedents across all rules. Results are shown for all sets of 3/5 classes: \emph{\textbf{De}sert road, \textbf{Dr}iveway, \textbf{F}orest, \textbf{H}ighway} and \emph{\textbf{S}treet}.}
	\begin{center}
		\footnotesize
		\begin{tabular}{| l | c | c | c | c | c | c | c | c | c | c | c | c | c | c | c |}
			\hline
			& \multicolumn{3}{c |}{Original $M$} & \multicolumn{3}{ c |}{Program $M^*$} & \multicolumn{3}{c |}{$M-M*$} & \multicolumn{3}{ c |}{$M^* stats$} \\
			\hline
			Classes & Train & Val. & Test & Tr. & Val. & Te. & Tr. & Val. & Te.& Vars & Rules & Size  \\
			\hline
			De,Dr,F & 98.5 & 88.5 & 90.2 & 83.5 & 79.7 & 81.6 & 15.0 & 8.8 & 8.6 & 50 & 31 & 171 \\
			De,Dr,H & 97.5 & 82.7 & 83.6 & 77.2 & 73.5 & 75.0 & 20.3 & 9.2 & 8.6 & 44 & 32 & 176 \\
			De,Dr,S & 99.6 & 92.9 & 93.3 & 78.7 & 74.9 & 76.1 & 20.9 & 18.0 & 17.2 & 44 &34  & 183 \\
			De,F,H & 95.0 & 80.8 & 81.5 & 85.0 & 80.7 & 81.4 & 10.0 & 0.1 & 0.1 & 48 & 36 & 196 \\
			\textbf{De,F,S} & 99.0 & 94.8 & 94.7 & 91.0 & \textbf{89.4} & 90.3 & 8.0 & 5.4 & 4.4 & 33 & 25 & 127 \\
			De,H,S & 97.7 & 84.9 & 86.2 & 80.6 & 78.0 & 78.7 & 17.1 & 6.9 & 7.5 & 42 & 36 & 194 \\
			Dr,F,H & 96.9 & 82.5 & 83.0 & 83.3 & 80.0 & 81.0 & 13.6 & 2.5 & 2.0 & 47 & 31 & 167 \\
			Dr,F,S & 97.9 & 89.7 & 90.9 & 73.4 & 68.9 & 69.6 & 24.5 & 20.8 & 21.3 & 47 & 33 & 181  \\
			Dr,H,S & 99.0 & 88.0 & 88.1 & 79.8 & 76.7 & 78.0 & 19.2 & 11.3 & 10.1 & 47 & 36 & 197 \\
			F,H,S & 97.7 & 86.9 & 87.1 & 73.8 & 71.2 & 71.6 & 23.9 & 15.7 & 15.5 & 56 & 34 & 185 \\
			\hline
		\end{tabular}
	\end{center}
	\label{table:netcompare}
\end{table}

\begin{figure}[t]
\centering
\includegraphics[width=120mm]{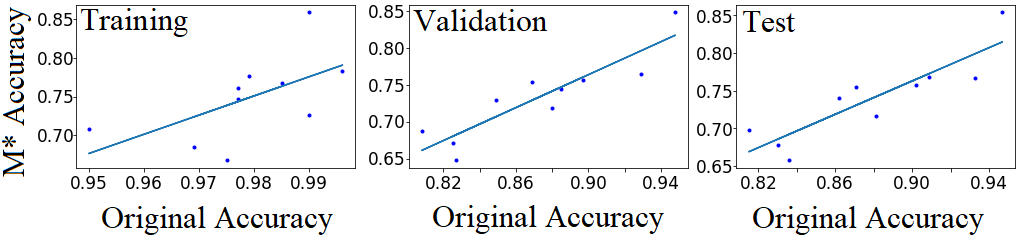} 
\caption{\emph{Original CNN accuracy} compared to accuracy of \emph{extracted model} $M*$}
\label{fig:acc_compare}
\end{figure}

We set $\alpha = 0.01$ so that branching stops if a branch represents less than $1\%$ of the samples represented by its parent node. We iterate the logical entry point $l^{LEP} \in [Conv8 \dots Conv13]$ and tree depth $d \in [1\dots5]$ and observe the effects on accuracy, fidelity and the size of the extracted program. Size is measured as the sum length (number of entecedents) of all rules and is our metric for interpretability on the basis that larger programs take longer to read and understand. In all figures we average over all results except the variable we iterate over.

\begin{figure}[t]
\centering
\includegraphics[width=120mm]{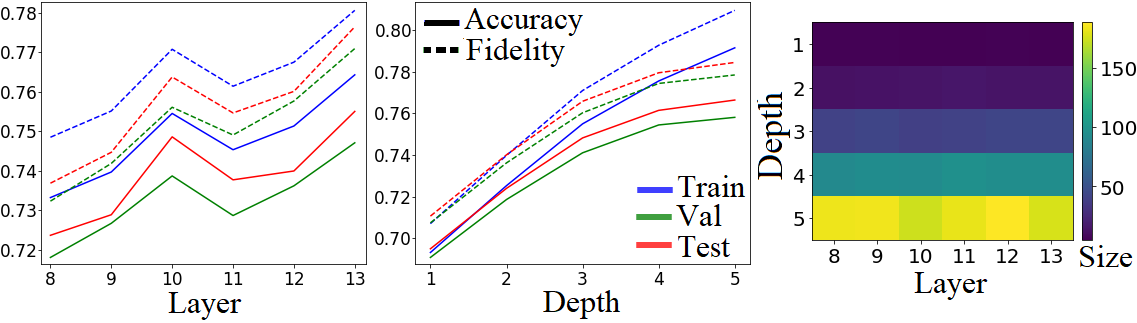} 
\caption{\emph{Accuracies}, \emph{fidelities} and program \emph{sizes} obtained from single-layer extraction.}
\label{fig:onelayer}
\end{figure}

Fig. \ref{fig:acc_compare} compares the average accuracy across all depths and layers for each class combination with the accuracy of the original model. The line of best fit shows that the accuracy of the extracted model is consistent with respect to that of the original model. In the validation set, accuracy drops by about 15\% in each case. However average validation accuracy drops by an average of 10\% for the optimal selection of depth and layer (table \ref{table:netcompare}). In summary, the loss in accuracy can be moderated by adjusting extraction parameters.

Fig. \ref{fig:onelayer} shows how accuracy, fidelity and program size are affected as we adjust the extraction layer and tree depth. Accuracy and fidelity both improve as tree depth (and therefore rule length) is increased, demonstrating that extraction benefits from quantising more kernels. However, the cost of this is a larger logic program. Accuracy and fidelity both show a general increase as rules are extracted from higher layers. This is to be expected since 1) deeper layers are known to represent more abstract and discriminative concepts, and 2) by discretising one layer we also discard all the information encoded in any following layers. However, there is a spike at layer 10. Layers 10 and 13 of VGG16 pass through max-pooling filters, suggesting that pooling before quantisation may also be beneficial to accuracy. The choice of extraction layer has small but negligible effect on program size. However in our case all extraction layers have 512 kernels and results may differ when extracting from smaller or larger layers.

Optimal validation accuracies were found for $conv13$ and a tree depth of $5$. Table \ref{table:netcompare} presents accuracies  for all class combinations based on this configuration. The best validation accuracy was found for \emph{Desert Road, Forest Road and Street} and table \ref{table:rules} shows example rules extracted for this case. Note that literals are composed of two letters because an alphabet of A-Z is insufficient for all 512 kernels, and they can be renamed in the kernel labelling stage anyway. We carry the optimal parameters and scenario forward for all further tests.

\begin{table}[b]
	\caption{6/25 Extracted rules for classes = \{desert road, forest road, street\}.}
	\begin{center}
		\footnotesize
\begin{tabular}{| c | l |}
	\hline
	1 & $LW \land \lnot SG \rightarrow street$ \\
	\hline
	7 & $CX \land \lnot LW \land NI \land PO \land \lnot SG \rightarrow street$ \\
	\hline
	10 & $\lnot CK \land DO \land \lnot HV \land JB \land NI \rightarrow forest$ \\
	\hline
	13 & $\lnot AC \land CK \land \lnot DO \land NI \land \lnot SG \rightarrow forest$ \\
	\hline
	17 & $\lnot AC \land \lnot DO \land \lnot JJ \land SG \rightarrow desert$ \\
	\hline
	25 & $AC \land \lnot DO \land ID \land SG \rightarrow desert$ \\
	\hline
\end{tabular}
	\end{center}
	\label{table:rules}
\end{table}

\subsection{Extraction from multiple layers}
\label{sect:exp_multi}

\begin{figure}[t]
\centering
\includegraphics[width=120mm]{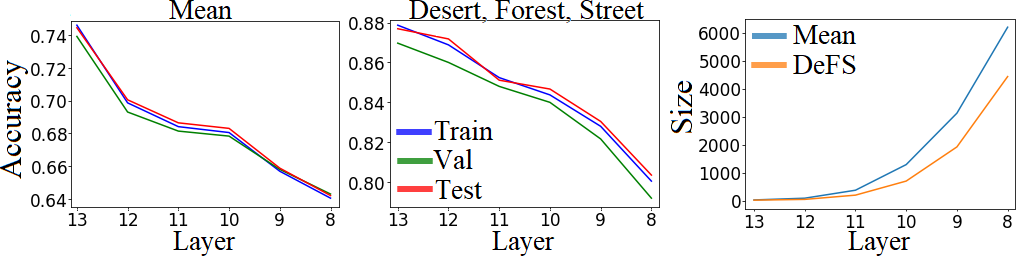} 
\caption{\emph{Accuracies}, \emph{fidelities} and program \emph{sizes} yielded from multi-layer extraction.}
\label{fig:mlayer}
\end{figure}

Given the higher complexity of extracting knowledge from multiple layers at once, we do not iterate different values for tree depth but fix it at 5. We also increase the value of $\alpha$ to $0.1$ to enforce stricter stopping conditions and prevent combinatorial explosion caused by observing relations between kernels at adjacent layers. Fig. \ref{fig:mlayer} shows the effect of incrementally adding layers starting from layer 13 only, then adding layer 12, and so on up to layer 8. Accuracy drops as more layers are added, presumably due to an increase in information loss as more and more kernels are quantised. Nonetheless accuracies are reasonable. However, the size of the logic program increases exponentially as more layers are added, emphasising the importance of adjusting $d$ and $\alpha$ to moderate this.

\subsection{Visualisation and labelling}
\label{sect:exp_vislab}

\begin{figure}[t!]
\centering
\includegraphics[width=98mm]{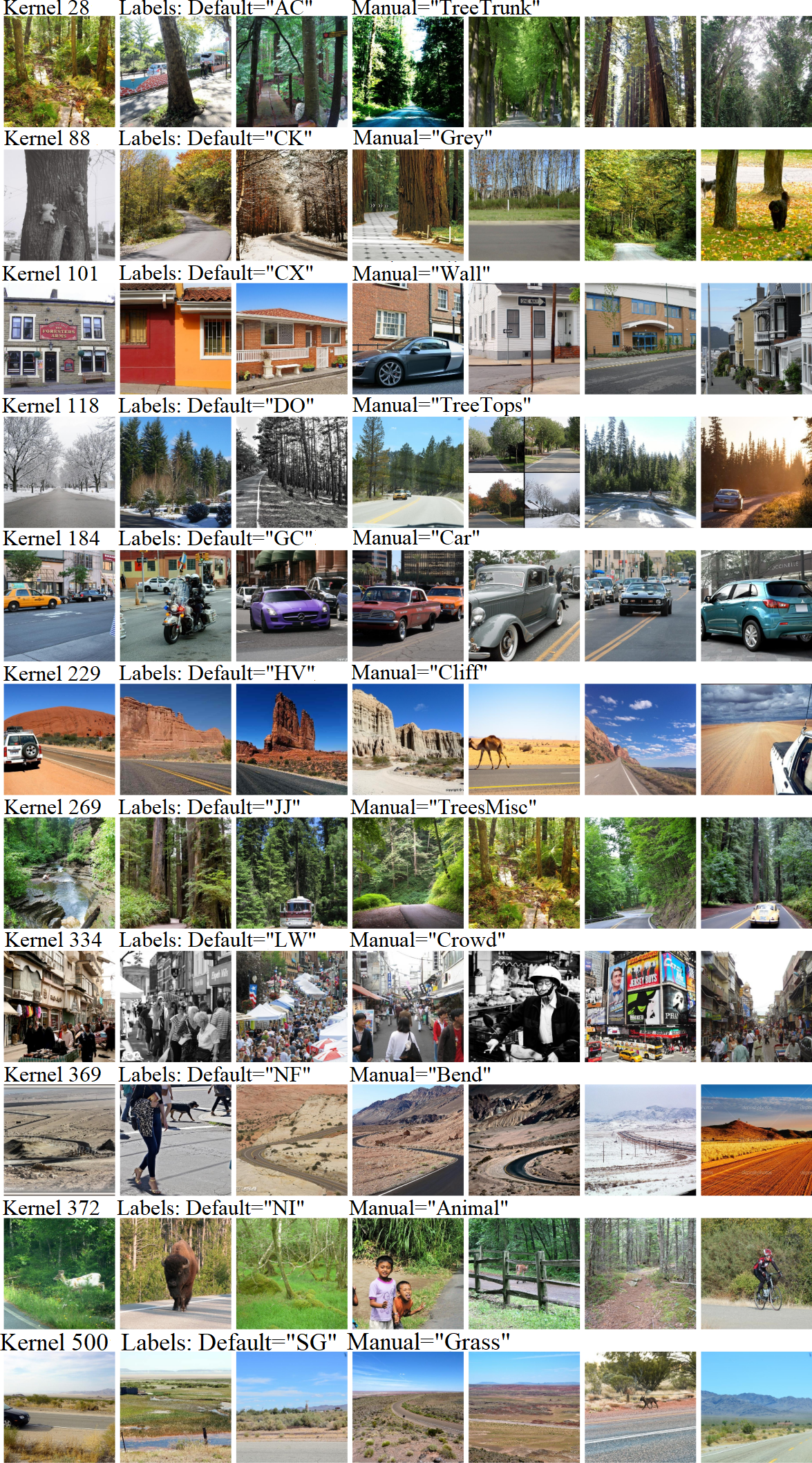} 
\caption{Kernel visualisations. All images from \emph{Places365} dataset \cite{zhou2017places}}
\label{fig:kvis}
\end{figure}

It can be difficult to know which kernels to examine when inspecting large CNNs such as VGG16 which has 512 in one layer. We have shown that using rule extraction this number can be reduced (the best case is 33 in table \ref{table:netcompare}). For now we assign labels manually with the intention of automating this process in future work. We label kernels represented by rules extracted for layer 13, with $m=10$. That is, we choose a kernel's label based on the 10 images that activate it most strongly with respect to \emph{l1} norm values obtained from a forward pass of the training set. Fig \ref{fig:kvis} presents 7/10 images\footnote{Limited space made it difficult to show all 10 without compromising clarity.} from $\hat{x}^{10}_{13,k}$ selected for 11 kernels. 

The kernels clearly partition classes into further sub-classes with noticable similarities within them, supporting the findings of Zhou et al. \cite{zhou2014object}. Images for kernels 101, 184 and 334 are all taken from the street class but focus on different things: 101 seems to focus on \emph{walls} mostly free from occlusion, 184 mostly on cars and 334  on \emph{crowds} of people. Kernels 229, 369 and 500 mostly respond to the desert class but again distinguish between different features: 229 responds strongly to \emph{cliffs} or mountains, 369 to \emph{bends} in roads and 500 to desert with \emph{grass} patches. The remaining kernels respond mostly to forest images but differences were less clear. Kernel 28 responds when \emph{tree trunks} are more visible and 118 when the \emph{tree tops} are visible. 88 was more difficult to choose a label for but we chose \emph{grey} due to significant grey regions in some images. 269 was also difficult to choose for. A common feature appears to be something in the centre of the image such as cars on two occasions, a lake on another and planks of wood in another. It may be that the kernel has picked up on a regularity not immediatly clear to human observers; an example of the need for a symbol for a possibly new concept to which we assign \emph{TreesMisc} as a surrogate.

However we label the kernels, the initial 3-class dataset does not have the labels necessary to distinguish between these sub-classes even though the CNN is capable of doing so to the extent that, as we show in sections \ref{sect:exp_single} and \ref{sect:exp_multi}, they can be quantised and included in rules that approximate the CNN's behaviour.

\subsection{Test images}
\label{sect:exp_rules}

\begin{figure}[t]
\centering
\includegraphics[width=120mm]{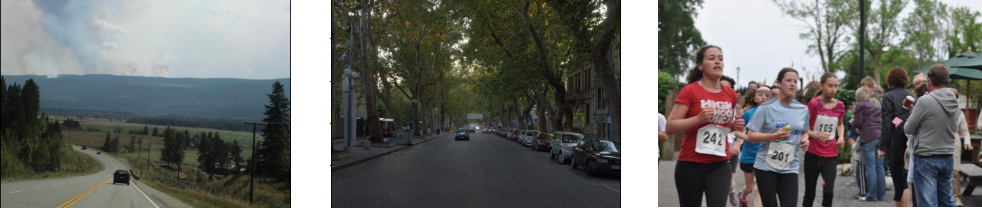} 
\caption{Misclassifications: a \emph{forest road} misclassified as a desert road, a \emph{street} misclassified as a forest, and a \emph{forest} misclassified as a street. Images from Places365 \cite{zhou2017places}.}
\label{fig:misclass}
\end{figure}

We now inspect some of the explanations given for classifications made on the test set, assigning the labels in fig. \ref{fig:kvis} to the rules in table \ref{table:rules}. Rule 1 translates as $Crowd \land \lnot Grass \rightarrow street$, i.e. ``if you encounter people and no grass then you are in a street''. Of course, in reality there are other situations where people may be found without grass, and some empty streets may have grass patches, so we as humans would not conclude we are in a street with this information alone. However, in this particular ``Desert, Forest or Street'' case on which the CNN was trained, one is significantly less likely to encounter people in the former two categories. Thus, this is enough information for the CNN to identify the location. Rule 7 translates as $Grey \land TreeTops \land Animal \rightarrow forest$. Animals may appear in streets, as would a grey surface, but when they appear together with trees it is more likely to be a forest. Rule 17 translates as $\lnot TreeTrunk \land \lnot TreeTops \land \lnot TreesMisc \land Grass \rightarrow desert$, i.e. ``if there is grass but no trees then it must be a desert''. Again, there are many places with grass but no trees (e.g. a field) but in this particular task the CNN has no other concept of grass without trees.

Fig. \ref{fig:misclass} shows three images for which both the original and approximated models made identical misclassifications. In the first example rule 17 misclassifies a forest road as a desert road. Although trees are present they are perhaps too few to activate the tree-related kernels, satisfying the negated tree-based antecedents. Grass by the road satisfies the other antecedent. In the second case rule 10 ($TreeTrunk \land \lnot Wall \land TreeTops  \land \lnot Crowd \land \lnot Animal \rightarrow forest$) confuses a street for a forest road as there are no animals in the street and many trees occlude the walls of the houses. The image from the forest set is misclassified as a street according to rule 1 as there are people and no grass.


\section{Discussion and future work}
\label{sect:discussion}

ERIC quantises the outputs of kernels in a CNN and relates these kernels to each other as logical rules that yield lower but reasonable accuracy. Our inspection of these kernels supported existing findings that most exhibit a strong response to sets of similar images with common semantic features \cite{zhou2014object}. We hope to automate the process of labelling these symbols in future work, likely integrating existing methods for mapping kernels or receptive fields to semantic concepts \cite{simon2014part,zhou2014object,zhang2017growing,zhang2018interpreting}. However these methods have finite sets of labels originally provided by humans with finite vocabularies. It may be that knowledge extraction methods will find new and important symbols for which we need to invent new words. ERIC provides a framework for discovering symbols that are important enough to distinguish between classes but for which no labels yet exist. 

Although the rules in section \ref{sect:exp_rules} are not paths of reasoning humans are likely to take, they nonetheless suffice to approximate the behaviour of the original CNN. It would be unreasonable for a human to assume they are in a street just because they see people and no grass, but for a CNN that has only seen streets, forests and desert roads, it is a reasonable assumption. Being able to explain how a machine `thinks' does not necessarily mean that it thinks like a human. 

An empirical comparison of performance of methods listed in the background also remains to be addressed in future work, but for now we comment on how they differ in terms of features. ERIC is a global explanation model that extracts rules in which antecedents are composed of positive and negative instances of quantised kernel activations, and is able to extract these rules from multiple convolutional layers. ERIC lacks some features that may be of benefit such as the ability to disentangle features and thus overcome assumptions regarding one-to-one relationships between kernels and concepts. However relationships defined using the disentanglement method do not include negated symbols as ERIC does. Both methods have potentially mutually beneficial features and adapting ERIC to disentangle representations would be an interesting future step.

Finally, although ERIC is not yet compatible with \emph{architectures} designed for explainability, we expect it would be compatible with weight matrices that have been \emph{trained} for explainability. We would like to test this hypothesis and use ERIC as a framework for assessing how this affects fidelity and explainability. 


\section{Conclusions}
\label{sect:conclusion}

We have shown that the behaviour of kernels across multiple convolutional layers can be approximated using a logic program, and the extracted program can be used as a framework in which we can begin to understand the behaviour of CNNs and how they think. More specifically, it can be used to identify kernels worthy of deeper inspection and their relationships with other kernels in the form of logical rules. Our own inspections show that the kernels in the last convolutional layer may be associated with concepts that are symbolic in the sense that they are visually distinct from those represented by other kernels. Some of these symbols were more interpretable from a human perspective than others. However regardless of what labels we assign, we have shown that these kernels can be used to construct symbolic rules that approximate the behaviour of the CNN to an accuracy that can be improved by adjusting rule length and the choice of layer or layers to extract from, at the cost of a larger and therefore less interpretable but nonetheless symbolic logic program. In the best case, we saw an average 10\% drop in accuracy compared with the original model.

\bibliographystyle{splncs}
\bibliography{0802}

\begin{thebibliography}{10}

\bibitem{ribeiro2016should}
Ribeiro, M.T., Singh, S., Guestrin, C.:
\newblock Why should {I} trust you?: Explaining the predictions of any
  classifier.
\newblock In: Proceedings of the 22nd ACM SIGKDD International Conference on
  Knowledge Discovery and Data Mining, ACM (2016)  1135--1144

\bibitem{gilpin2018explaining}
Gilpin, L.H., Bau, D., Yuan, B.Z., Bajwa, A., Specter, M., Kagal, L.:
\newblock Explaining explanations: An overview of interpretability of machine
  learning.
\newblock In: 2018 IEEE 5th International Conference on Data Science and
  Advanced Analytics (DSAA), IEEE (2018)  80--89

\bibitem{guidotti2018survey}
Guidotti, R., Monreale, A., Ruggieri, S., Turini, F., Giannotti, F., Pedreschi,
  D.:
\newblock A survey of methods for explaining black box models.
\newblock ACM computing surveys (CSUR) \textbf{51} (2018) ~93

\bibitem{andrews1995survey}
Andrews, R., Diederich, J., Tickle, A.B.:
\newblock Survey and critique of techniques for extracting rules from trained
  artificial neural networks.
\newblock Knowledge-based systems \textbf{8} (1995)  373--389

\bibitem{jacobsson2005rule}
Jacobsson, H.:
\newblock Rule extraction from recurrent neural networks: A taxonomy and
  review.
\newblock Neural Computation \textbf{17} (2005)  1223--1263

\bibitem{townsend2019extracting}
Townsend, J., Chaton, T., Monteiro, J.M.:
\newblock Extracting relational explanations from deep neural networks: A
  survey from a neural-symbolic perspective.
\newblock IEEE transactions on neural networks and learning systems (2019)

\bibitem{zhang2018visual}
Zhang, Q., Zhu, S.:
\newblock Visual interpretability for deep learning: a survey.
\newblock Frontiers of Information Technology \& Electronic Engineering
  \textbf{19} (2018)  27--39

\bibitem{lamb2020graph}
Lamb, L., Garcez, A., Gori, M., Prates, M., Avelar, P., Vardi, M.:
\newblock Graph neural networks meet neural-symbolic computing: A survey and
  perspective.
\newblock arXiv preprint arXiv:2003.00330 (2020)

\bibitem{simonyan2013deep}
Simonyan, K., Vedaldi, A., Zisserman, A.:
\newblock Deep inside convolutional networks: Visualising image classification
  models and saliency maps.
\newblock arXiv preprint arXiv:1312.6034 (2013)

\bibitem{zeiler2014visualizing}
Zeiler, M.D., Fergus, R.:
\newblock Visualizing and understanding convolutional networks.
\newblock In: European conference on computer vision, Springer (2014)  818--833

\bibitem{springenberg2014striving}
Springenberg, J.T., Dosovitskiy, A., Brox, T., Riedmiller, M.:
\newblock Striving for simplicity: The all convolutional net.
\newblock arXiv preprint arXiv:1412.6806 (2014)

\bibitem{bojarski2016visualbackprop}
Bojarski, M., Choromanska, A., Choromanski, K., Firner, B., Jackel, L., Muller,
  U., Zieba, K.:
\newblock Visualbackprop: efficient visualization of cnns.
\newblock arXiv preprint arXiv:1611.05418 (2016)

\bibitem{bach2015pixel}
Bach, S., Binder, A., Montavon, G., Klauschen, F., M{\"u}ller, K., Samek, W.:
\newblock On pixel-wise explanations for non-linear classifier decisions by
  layer-wise relevance propagation.
\newblock PloS one \textbf{10} (2015)  e0130140

\bibitem{samek2017evaluating}
Samek, W., Binder, A., Montavon, G., Lapuschkin, S., M{\"u}ller, K.:
\newblock Evaluating the visualization of what a deep neural network has
  learned.
\newblock IEEE transactions on neural networks and learning systems \textbf{28}
  (2017)  2660--2673

\bibitem{shrikumar2017learning}
Shrikumar, A., Greenside, P., Kundaje, A.:
\newblock Learning important features through propagating activation
  differences.
\newblock arXiv preprint arXiv:1704.02685 (2017)

\bibitem{frosst2017distilling}
Frosst, N., Hinton, G.:
\newblock Distilling a neural network into a soft decision tree.
\newblock arXiv preprint arXiv:1711.09784 (2017)

\bibitem{chen2019looks}
Chen, C., Li, O., Tao, D., Barnett, A., Rudin, C., Su, J.K.:
\newblock This looks like that: deep learning for interpretable image
  recognition.
\newblock In: Advances in Neural Information Processing Systems. (2019)
  8930--8941

\bibitem{bologna2020two}
Bologna, G., Fossati, S.:
\newblock A two-step rule-extraction technique for a cnn.
\newblock Electronics \textbf{9} (2020)  990

\bibitem{zhang2017growing}
Zhang, Q., Cao, R., Wu, Y.N., Zhu, S.:
\newblock Growing interpretable part graphs on convnets via multi-shot
  learning.
\newblock In: Thirty-First AAAI Conference on Artificial Intelligence. (2017)

\bibitem{zhang2018interpreting}
Zhang, Q., Cao, R., Shi, F., Wu, Y.N., Zhu, S.:
\newblock Interpreting {CNN} knowledge via an explanatory graph.
\newblock In: Thirty-Second AAAI Conference on Artificial Intelligence. (2018)

\bibitem{zhang2019interpreting}
Zhang, Q., Yang, Y., Ma, H., Wu, Y.N.:
\newblock Interpreting cnns via decision trees.
\newblock In: Proceedings of the IEEE Conference on Computer Vision and Pattern
  Recognition. (2019)  6261--6270

\bibitem{zhou2014object}
Zhou, B., Khosla, A., Lapedriza, A., Oliva, A., Torralba, A.:
\newblock Object detectors emerge in deep scene {CNN}s.
\newblock arXiv preprint arXiv:1412.6856 (2014)

\bibitem{zhang2018interpretable}
Zhang, Q., Nian~Wu, Y., Zhu, S.:
\newblock Interpretable convolutional neural networks.
\newblock In: Proceedings of the IEEE Conference on Computer Vision and Pattern
  Recognition. (2018)  8827--8836

\bibitem{percy2016need}
Percy, C., d'Avila Garcez, A.S., Dragicevic, S., Fran{\c{c}}a, M.V., Slabaugh,
  G.G., Weyde, T.:
\newblock The need for knowledge extraction: Understanding harmful gambling
  behavior with neural networks.
\newblock Frontiers in Artificial Intelligence and Applications \textbf{285}
  (2016)  974--981

\bibitem{ribeiro2018anchors}
Ribeiro, M.T., Singh, S., Guestrin, C.:
\newblock Anchors: High-precision model-agnostic explanations.
\newblock In: AAAI Conference on Artificial Intelligence. (2018)

\bibitem{zilke2016deepred}
Zilke, J.R., Menc{\'\i}a, E.L., Janssen, F.:
\newblock Deepred--rule extraction from deep neural networks.
\newblock In: International Conference on Discovery Science, Springer (2016)
  457--473

\bibitem{schaaf2019enhancing}
Schaaf, N., Huber, M.F.:
\newblock Enhancing decision tree based interpretation of deep neural networks
  through l1-orthogonal regularization.
\newblock arXiv preprint arXiv:1904.05394 (2019)

\bibitem{nguyen2020towards}
Nguyen, T.D., Kasmarik, K.E., Abbass, H.A.:
\newblock Towards interpretable deep neural networks: An exact transformation
  to multi-class multivariate decision trees.
\newblock arXiv preprint arXiv:2003.04675 (2020)

\bibitem{murdoch2017automatic}
Murdoch, W.J., Szlam, A.:
\newblock Automatic rule extraction from long short term memory networks.
\newblock arXiv preprint arXiv:1702.02540 (2017)

\bibitem{tran2016deep}
Tran, S.N., d'Avila Garcez, A.S.:
\newblock Deep logic networks: Inserting and extracting knowledge from deep
  belief networks.
\newblock IEEE transactions on neural networks and learning systems (2016)

\bibitem{zhou2016learning}
Zhou, B., Khosla, A., Lapedriza, A., Oliva, A., Torralba, A.:
\newblock Learning deep features for discriminative localization.
\newblock In: Computer Vision and Pattern Recognition (CVPR), 2016 IEEE
  Conference on, IEEE (2016)  2921--2929

\bibitem{denil2014extraction}
Denil, M., Demiraj, A., De~Freitas, N.:
\newblock Extraction of salient sentences from labelled documents.
\newblock arXiv preprint arXiv:1412.6815 (2014)

\bibitem{selvaraju2017grad}
Selvaraju, R.R., Cogswell, M., Das, A., Vedantam, R., Parikh, D., Batra, D.:
\newblock Grad-cam: Visual explanations from deep networks via gradient-based
  localization.
\newblock In: Proceedings of the IEEE international conference on computer
  vision. (2017)  618--626

\bibitem{simon2014part}
Simon, M., Rodner, E., Denzler, J.:
\newblock Part detector discovery in deep convolutional neural networks.
\newblock In: Asian Conference on Computer Vision, Springer (2014)  162--177

\bibitem{xie2017relating}
Xie, N., Sarker, M.K., Doran, D., Hitzler, P., Raymer, M.:
\newblock Relating input concepts to convolutional neural network decisions.
\newblock arXiv preprint arXiv:1711.08006 (2017)

\bibitem{zhou2017places}
Zhou, B., Lapedriza, A., Khosla, A., Oliva, A., Torralba, A.:
\newblock Places: A 10 million image database for scene recognition.
\newblock IEEE Transactions on Pattern Analysis and Machine Intelligence (2017)

\bibitem{quinlan1993c4}
Quinlan, J.R.:
\newblock C4.5: Programming for machine learning.
\newblock The Morgan Kaufmann Series in Machine Learning, San Mateo, CA: Morgan
  Kaufmann \textbf{38} (1993) ~48

\bibitem{chollet2015keras}
Chollet, F.,  et~al.:
\newblock Keras.
\newblock \url{https://github.com/fchollet/keras} (2015)

\end{thebibliography}

\end{document}